\documentclass[runningheads]{llncs}
\usepackage{graphicx}
\usepackage{mathtools} % amsmath with fixes and additions
\usepackage{booktabs} % commands to create good-looking tables
\usepackage{amsmath}
\usepackage{amssymb}
\usepackage{mathtools}
\usepackage{xcolor}
\usepackage{listings}
\usepackage{multirow}
\usepackage[ruled,vlined,linesnumbered]{algorithm2e}

\newenvironment{sloppypar*}
 {\sloppy\ignorespaces}
 {\par}

\usepackage{xspace}

\newcommand{\ie}{\textit{i.e.}\xspace}

\newcommand{\eg}{\textit{e.g.}\xspace}
\newcommand{\etc}{\textit{etc.}\xspace}

\DeclareGraphicsExtensions{.pdf,.png,.jpg,.bmp}

\begin{document}
\title{Feature Importance versus Feature Influence and What It Signifies for Explainable AI\thanks{The work is partially supported by the Wallenberg AI, Autonomous Systems and Software Program (WASP) funded by the Knut and Alice Wallenberg Foundation.}}

\titlerunning{Feature Importance versus Feature Influence}
\author{Kary Fr\"{a}mling\inst{1,2}\orcidID{0000-0002-8078-5172}}
\authorrunning{K. Fr\"{a}mling}
\institute{Ume\r{a} University, 901 87 Ume\r{a}, Sweden \and Aalto University, 02150 Espoo, Finland \\
\email{kary.framling@cs.umu.se}\\
\url{https://www.umu.se/personal/kary-framling/}}
\maketitle              % typeset the header of the contribution
\begin{abstract}
When used in the context of decision theory, \textit{feature importance} expresses how much changing the value of a feature can change the model outcome (or the \textit{utility} of the outcome), compared to other features. Feature importance should not be confused with the \textit{feature influence} used by most state-of-the-art post-hoc Explainable AI methods. Contrary to feature importance, feature influence is measured against a \textit{reference level} or \textit{baseline}. The Contextual Importance and Utility (CIU) method provides a unified definition of global and local feature importance that is applicable also for post-hoc explanations, where the \textit{value utility} concept provides instance-level assessment of how favorable or not a feature value is for the outcome. The paper shows how CIU can be applied to both global and local explainability, assesses the fidelity and stability of different methods, and shows how explanations that use contextual importance and contextual utility can provide more expressive and flexible explanations than when using influence only. 

\end{abstract}
\section{Introduction}

Explainable Artificial Intelligence (XAI) is probably as old as AI itself and papers even from the 1970's such as \cite{SHORTLIFFE1975303} can still give valuable insight to XAI researchers. A relatively small but active XAI community existed in the 1990's, which tended to focus on building rule-based surrogate models of trained neural networks \cite{AndrewsEtAl_1995}. The \textit{Contextual Importance and Utility (CIU)} method was presented in \cite{FramlingThesis_1996,FramlingAISB_1996} at the same epoch and proposed a different approach, where only the outcome of the black-box model for a specific instance was justified and explained, which is nowadays often called \textit{post-hoc explanation}. However, it seems like CIU was forgotten since then.
%CIU has been cited only in a few papers but been set aside due to claims such as ``importance and utility can be efficiently computed only in a special INKA network''  \cite{RobnikSikonja_2008,RobnikSikonja_Neurocomputing_2012}. 
%and ``CIU can be misleading when F is not monotonous'' and ``this method is based on extremums variable and thus very sensitive to noise'' \cite{LemaireEtAl_2008}. 
%In practice, CIU is entirely model-agnostic as shown recently \eg by \cite{EXTRAAMAS_2021_CIU_Framling}. This paper evaluates the computation efficiency of CIU in general and compared to current state-of-the-art model-agnostic post-hoc XAI methods, namely the family of \textit{additive feature attribution (AFA)} methods \cite{NIPS2017_Lundberg_XAI}. 

This paper revisits the theoretical foundations of CIU and the core differences between CIU and current state-of-the-art methods. Notably, we show that ``contextual importance'' is compatible with the notion of ``global feature importance'' but extends it to the case of instance-level post-hoc explanation. We also show that influence values produced by so called Additive Feature Attribution (AFA) methods \cite{NIPS2017_Lundberg_XAI} are not compatible with global feature importance. %However, influence values are compatible with the notion of contextual influence, which is calculated from contextual importance and contextual utility. 
Finally, we compare the methods experimentally using general-purpose criteria.

After this Introduction, Section \ref{Sec:Methods} goes through the theory of global feature importance and AFA methods, followed by a presentation of CIU and new theory in Section~\ref{Sec:CIU}. Section \ref{Sec:Experiments} experimentally compares CIU with the methods presented in Section \ref{Sec:Methods}, followed by Conclusions.  

\section{Background}\label{Sec:Methods}

When explaining the outcome of a model $f$, we are interested in how each feature (or possibly groups of features) affects the prediction of the instance being explained. For a linear model it is easy to identify the importance and influence of each feature, where the prediction for an instance $x$ is: 

\begin{equation}
\label{Eqn:LinearFunction}
    f(x)=w_0+w_{1}x_{1}+\dots+w_{N}x_{N}.
\end{equation}

%where $x$ is the instance for which we want to explain the outcome $f(x)$. 
Each $x_{i}$ is a feature value, with $i = 1,\dots,N$ and $w_{i}$ is the weight associated with the feature $i$. Such linear models are considered understandable to humans and are therefore often used as so-called surrogate models $g$ for explaining the outcome also for non-linear models, such as those produced by machine learning methods. 

However, there are fundamental differences in how different XAI methods define \textit{feature importance} in Equation~\ref{Eqn:LinearFunction}, versus \textit{feature influence}. In decision theory and most other contexts, the feature importance for a linear model like Equation~\ref{Eqn:LinearFunction} is considered to be $w_{i}$ for feature $i$ \cite{Fishburn1990,keeney_raiffa_1993}. This is the definition that we adopt also in this paper. In \textit{multi-attribute utility theory (MAUT)}, $x_{i}$ is replaced by a \textit{utility value} given by a \textit{utility function} $u_{i}(x_{i})$ \cite{Dyer2005}.

%Other concepts are also used in the sense of importance or influence even within the same source, such as effect, relevance or contribution. 
On the other hand, the state-of-the-art XAI method Shapley value calculates an influence value $\phi_{i}$ for each feature. For a linear model as in Equation~\ref{Eqn:LinearFunction}, the Shapley value $\phi_{i}$ of the $i$-th feature on the prediction $f(x)$ is:

\begin{equation}\label{Eq:Influence}
    \phi_{i}(x)=w_{i}x_{i}-E(w_{i}X_{i})=w_{i}x_{i}-w_{i}E(X_{i}),
\end{equation}

where $E(w_{i}X_{i})$ is the mean effect estimate for feature $i$  \cite{StrumbeljKononenko_2014}. This influence value $phi_{i}(x)$ depends on the instance $x$ and is clearly not the same as the importance $w_{i}$. %The mean effect can be estimated in different ways by different methods and is usually called the \textit{baseline} or \textit{reference level}. 

\noindent The concepts of importance, utility, utility function and influence can be illustrated by an example of how the weighted average grade of a university student is calculated. The weight $w_{i}$ of each course is the number of credits that can be obtained for the course and corresponds to the importance of the course. The utility value $u_{i}$ of each course is the obtained grade in percent, \ie normalized into range $[0,1]$ by a utility function $u_{i}(x_{i})$, where $x_{i}$ is the original grade. The result $u=\sum_{i=1}^{N}w_{i}u_{i}(x_{i})$ is normalized to the range $[0,1]$ by dividing the weights $w_{i}$ by the sum of all weights. Finally, the weighted average grade is scaled from the range $[0,1]$ to the desired output range using the utility function $u(y)$, \eg $[0,1] \mapsto [4,10]$. The desired output range tends to vary between schools, universities and countries.  

The question is then: ``where do we have `influence' in the calculation of weighted average grade?''. The answer is ``nowhere'', unless we introduce a baseline or reference level, which would normally be the average grade for the group of reference students that we want to compare with. The $\phi_{i}$ values from Equation~\ref{Eq:Influence} would then indicate which courses had a negative or positive influence compared to the average of the reference population (\eg students following the same cursus the same year), and a magnitude for that influence.

\subsection{Global Feature Importance}
\label{Sec:GlobalFeatureImportance}

In a machine learning context, global feature importance describes how much covariates contribute to a prediction model’s accuracy, which differs from how decision theory defines it. However, both definitions are similar in practice as long as we only consider linear models, which is shown by the results in Section~\ref{Sec:Experiments}. 

Estimating global feature importance can be done in many ways, as described \eg in~\cite{FisherEtAl_JMLR:v20:18-760}. One commonly used method is the permutation-based feature importance (PFI) approach proposed in~\cite{Breiman_2001_RandomForests}, which works by calculating the increase of the model’s prediction error after permuting the feature values. A feature is ``important'' if permuting its values increases the model error, because the model relied on the feature for the prediction. A feature is ``unimportant'' if permuting its values keeps the model error unchanged, because the model ignored the feature for the prediction. 

% cite{FisherEtAl_JMLR:v20:18-760}: https://www.jmlr.org/papers/volume20/18-760/18-760.pdf

%That is only applicable for for tree-based models. 
% https://www.nature.com/articles/s42256-019-0138-9#citeas
%https://arxiv.org/pdf/1905.04610.pdf
%\textcolor{blue}{``Calculating low variance estimates of the Shapley values for the results in this paper would be intractable''. ``For moderate sized models, TreeExplainer is several orders of magnitude faster than model-agnostic alternatives, and has zero estimation variability''. In Figure 4A, they show "Global Feature Importance" as $mean(|SHAP value|)$}

%\url{http://uc-r.github.io/iml-pkg#global}

\subsection{Additive Feature Attribution Methods}
\label{Sec:AFA_methods}

The concept of AFA methods was introduced in \cite{NIPS2017_Lundberg_XAI}, where a set of XAI methods were identified that belong to this family. AFA methods use an \textit{explanation model} $g$ that is an interpretable approximation of the original model $f$, according to the following definition. 

\begin{definition} \textbf{AFA methods} have an explanation model that is a linear function of binary variables:
\begin{equation}
g(z')=\phi_{0}+\sum_{i=1}^{M}\phi_{i}z'_{i}, 
\label{Eq:AdditiveFeatureAttribution}
\end{equation}

where $z'\in \{0,1\}^M$, $M\leq N$ is the number of simplified input features, and $\phi \in \mathbb{R}$. 
\label{Def:AdditiveFeatureAttribution}
\end{definition}

\noindent According to \cite{LundbergEtAl_TreeSHAP_2018}, 
Shapley values represent the only possible method in the broad class of AFA methods that will simultaneously satisfy three important properties: \textit{local accuracy}, \textit{consistency}, and \textit{missingness}.
Local accuracy states that when approximating the original model $f$ for a specific input $x$, the explanation’s influence values should sum up to the difference $f(x)-\phi_{0}$.

%Methods with explanation models $g$ matching this definition attribute an influence $\phi_{i}$ to each input feature. 

%\textcolor{blue}{Alternatively, use description in https://christophm.github.io/interpretable-ml-book/shapley.html}

%\textcolor{blue}{When applying this to Shapley values, it's only valid for linear models. In all (?) other cases, using the average value $E(X_{i})$ as the so-called baseline is not necessarily an appropriate choice [LOADS OF REFERENCES HERE, see e.g. https://arxiv.org/pdf/2105.10719.pdf].}

%The idea behind Shapley values is to assess every combination of predictors to determine each predictor's impact. 
The Shapley value is a solution concept that assigns a pay-off to each agent according to their marginal contribution \cite{shapley:book1952}. Focusing on feature $i$, the Shapley value approach will test the accuracy of every combination of features not including feature $i$ and then test how adding $x_{i}$ to each combination improves the accuracy. Computing Shapley values is computationally expensive so most model-agnostic implementations only estimate approximate Shapley values, such as \textit{Kernel SHAP} \cite{NIPS2017_Lundberg_XAI}. Kernel SHAP is essentially an adaptation of another AFA method, the Local Interpretable Model-agnostic Explanations (LIME) method \cite{ribeiro2016should}, to estimate Shapley values. There are also model-specific methods for estimating Shapley values, such as Deep SHAP and Tree SHAP \cite{LundbergEtAl_TreeSHAP_2018}. 

Shapley values are ``local'' in the sense that they are calculated for a specific instance $x$. However, in \cite{LundbergEtAl_TreeSHAP_2018} it was suggested that $mean(|\phi_{i}|)$ could be used as a global feature importance estimate, when calculated over the entire training set and where influence values $\phi_{i}$ are Shapley values. They compare their approach with three well-known global feature importance methods and conclude that $mean(|\phi_{i}|)$ provides a better estimate of global feature importance. However, it can be questioned whether the use of influence values for estimating importance values is reasonable. As shown in this paper, contextual importance is mathematically similar to global feature importance. Different approaches for estimating global feature importance are used in Section~\ref{Sec:Experiments} that highlight these differences.

Shapley value and LIME are presumably the most used methods for the moment within the category of model-agnostic post-hoc XAI. Shapley value seem to become the dominating method due to its mathematical properties, as explained above. However, for instance \cite{kumar2020problems} recently pointed out several mathematical and human-centric issues with the use of Shapley value for XAI purposes. The mathematical issues concern how influence values are estimated and whether to use conditional or interventional distributions. %Even more importantly, the ``additivity'' property of Shapley values imposes severe limitations because constraining influence values has implications for what kinds of models can be explained intuitively by the Shapley value. 
Even in simple cases, the Shapley value is conceptually limited for non-additive models. Human-centric issues are mainly that Shapley value supports contrastive explanations only in comparison with the mean influence $E(w_{i}X_{i})$ and are difficult to use for producing other kinds of counterfactual explanations. It is also noted that even when an individual lacks a correct
mental model of the meaning of Shapley values, the explainee may use them to justify their evaluation anyways, whether or not this analysis is well-founded. In general, it is a problem if the explainability method is a black-box itself because then the explainee tends to interpret the explanations according to assumptions that might be false. 

%\cite{EXTRAAMAS_2021_CIU_Framling} focus on the conceptual difference between influence and importance and partially similar problems with influence values as the ones identified by \cite{kumar2020problems}. They suggest that CIU avoids many of those problems.  

%Shapley values also suffer of OOD (Out Of Distribution) challenge when estimating them. Since the Monte-Carlo sampling is done for all $x_{i}$ except the studied one, the OOD problem is presumably much bigger for Shapley values than for CIU (except when moving to intermediate concepts??).  

%Kolla denna: http://proceedings.mlr.press/v119/kumar20e/kumar20e.pdf "Problems with Shapley-value-based explanations as feature importance measures". They also use "influence" to describe $\phi$ values. \cite{kumar2020problems}.

\section{Contextual Importance and Utility}\label{Sec:CIU}

CIU is inspired from MAUT, which addresses the question of how humans can express their preferences and how they can be modelled mathematically in order to build human-understandable decision support systems. CIU uses core MAUT concepts of feature importance and utility value and specifies how they can be estimated for any model $f(x)$ and a specific instance or \textit{Context} $x$. In MAUT, an N-attribute utility function is a weighted sum: 

\begin{equation}
u(y)=u(x_{1},\dots ,x_{N})=\sum _{i=1}^{N}{w_{i}u_{i}(x_{i})},
\label{Eq:n_attribute_utility_function}
\end{equation}

\begin{sloppypar*}
where $u_{i}(x_{i})$ are the utility functions that correspond to the features $x_{1},\dots,x_{N}$. $u_{i}(x_{i})$ are constrained to the range [0,1], as well as $u(y)$ through the positive weights $w_{i}$. %Equation~\ref{Eq:n_attribute_utility_function} is identical to Equation~\ref{Eqn:LinearFunction} when replacing $w_{i}$ with $k_{i}$ and $x_{i}$ with $u_{i}(x_{i})$, which gives us the function:
%\begin{equation}
%\label{Eqn:LinearUtilityFunction}
%    u(x)=w_{1}u_{1}(x_{1})+\dots+w_{N}u_{N}(x_{N})
%\end{equation}
\end{sloppypar*}

When studying a general (presumably non-linear) model $f(x)$, there are no known $w_{i}$ values, so we need a way to estimate those values. 

\paragraph{Contextual Importance (CI).} 

CI takes the range of variation of the output as the importance value to estimate, which we can do by observing changes in output values when modifying input values of the feature $i$ and keeping the values of the other features at those given by the studied instance $x$. This principle can be extended to more than one feature and we will use the notion of input index set $\{i\}$ in the CIU equations that follow. For an input index set $\{i\}$, the values of the features $\neg \{i\}$ are defined by the instance $x$. We extend the definition further to the importance of $\{i\}$ versus another set of inputs $\{I\}$, where $\{i\}\subseteq \{I\} \subseteq {1,\dots,N}$. This gives us an estimation of the range $[umin_{j,\{i\},\{I\}}(x),umax_{j,\{i\},\{I\}}(x)]$, where $j$ is the index of the model output to explain. CI is defined as:

\begin{equation}
    \label{Eq:CI_definition}
    \omega_{j,\{i\},\{I\}}(x)=\frac{umax_{j,\{i\}}(x)-umin_{j,\{i\}}(x)}{umax_{j,\{I\}}(x)-umin_{j,\{I\}}(x)},
\end{equation}

\begin{sloppypar*}
where we use the symbol $\omega$ for CI. If the model $f(x)$ is linear, then $\omega_{j,\{i\},\{I\}}(x)$ is identical for all/any instance $x$ and is therefore also the global feature importance. If the model is non-linear, then $\omega_{j,\{i\},\{I\}}(x)$ depends on the instance $x$ and is ``local'' or \textit{contextual}.
\end{sloppypar*}

Since the model $f$ produces actual output values rather than utility values, the utility values $umin_{j}$ and $umax_{j}$ in Equation~\ref{Eq:CI_definition} have to be mapped to actual output values $y_{j}=f(x)$. If $f$ is a classification model, then the outputs $y_{j}$ are typically estimated probabilities for the corresponding class, so the output value is also the utility value, \ie $u_{j}(y_{j})=y_{j}$. For many (or most) regression tasks, a utility function of the form $u_{j}(y_{j})=Ay_{j}+b$ is applicable (and also applies to $u_{j}(y_{j})=y_{j}$). When the utility function is of form $u_{j}(y_{j})=Ay_{j}+b$, then CI can be calculated as:

\begin{equation} \label{Eq:OriginalCI}
\omega_{j,\{i\},\{I\}}(x)= \frac{ymax_{j,\{i\}}(x)-ymin_{j,\{i\}}(x)}{ ymax_{j,\{I\}}(x)-ymin_{j,\{I\}}(x)}, 
\end{equation}

where $ymin_{j}()$ and $ymax_{j}()$ are the minimal and maximal $y_{j}$ values observed for output $j$. %Equation~\ref{Eq:OriginalCI} is identical to the original CI definition by~\cite{FramlingThesis_1996,FramlingAISB_1996}.

% It is worth pointing out that the sum $\sum_{i=1}^{N}{\omega_{i}}$ can in principle be as big as $N$ when the model $f$ is non-linear. For instance, the binary OR function of two variables and the instance (0,0) will give $\omega=1$ for both variables. Shapley values would here have $E(w_{i}X_{i})=0.75$ and actual value zero, so $\phi=-0.75/2=-0.375$ for both variables. Despite such dependent variables, the joint importance $w_{\{i\},\{I\}}$ can not be greater than one by definition:

% \begin{definition}[Joint importance of all features]
% \label{Def:JointImpAllFeatures}
% The joint importance of all features is one, \ie when the index set $\{i\}=\{1,\dots,N\}$, then $CI_{\{i\}} = 1$.
% \end{definition}

% Since $\omega > 0$ by definition, we get:

% \begin{theorem}[Maximal range of Contextual Importance]
% $\omega_{j,\{i\}}(x) \in [0,1]$ for any set of features $\{i\}$. 
% \label{The:CI_Range}
% \end{theorem}

\paragraph{Contextual Utility (CU).} CU corresponds to the factor $u_{i}(x_{i})$ in Equation \ref{Eq:n_attribute_utility_function}. CU expresses to what extent the current values of features in $\{i\}$ contribute to obtaining a high output utility $u_{j}$. CU is defined as

\begin{equation}
CU_{j,\{i\}}(x)=\frac{u_{j}(x)-umin_{j,\{i\}}(x)}{umax_{j,\{i\}}(x)-umin_{j,\{i\}}(x)}. 
\label{Eq:CU}
\end{equation}

When $u_{j}(y_{j})=Ay_{j}+b$, this can be written as:  

\begin{equation}\label{Eq:CU_y}
CU_{j,\{i\}}(x)=\left|\frac{ y_{j}(x)-yumin_{j,\{i\}}(x)}{ymax_{j,\{i\}}(x)-ymin_{j,\{i\}}(x)}\right|, 
\end{equation}

where $yumin=ymin$ if $A$ is positive and $yumin=ymax$ if $A$ is negative. %This definition of CU differs from CI definitions in \cite{FramlingThesis_1996,Framling_XAI_WS_AAAI_2021} by handling negative $A$ values correctly. 

\paragraph{Contextual Influence.}\label{Sec:ContextualInfluence}

Contextual Influence defines feature influence in a similar way as in Equation~\ref{Eq:Influence} but using $w_{i}u_{i}(x_{i})$ instead of $w_{i}x_{i}$. This gives us 

\begin{equation}\label{Eq:CIU_Influence}
    \phi_{i}(x)=\omega_{i}(x)(u_{i}(x_{i}) - E(U(x_{i})),
\end{equation}

where $E(U(x_{i})$ is the expected utility value for feature $i$. Since utility $u \in [0,1]$ for all features, it intuitively makes sense to use the average utility value $0.5$ as a constant baseline for all features, even though it can be any value in the range $[0,1]$ that makes sense for the application. For consistency, this constant is called $\phi_{0}$ as in Equation~\ref{Eq:AdditiveFeatureAttribution}. When including subset indices for CI and CU, we get the following equation for contextual influence:

\begin{equation}\label{Eq:CIU_InfluenceFinal}
    \phi_{j,\{i\},\{I\}}(x)=\omega_{j,\{i\},\{I\}}(x)(CU_{j,\{i\}}(x) - \phi_{0}).
\end{equation}

\noindent Contextual influence makes it possible to produce influence-based explanations like for AFA methods, in addition to the explanations based on CI and CU. 

It should be emphasized that the baseline $\phi_{0}$ of contextual influence has a constant and semantic meaning, \ie ``averagely good/bad'', ``averagely typical'' \etc, that presumably makes sense to humans when used in explanations. It is also entirely data- and model-agnostic, which makes it different from the AFA baseline in Equation~\ref{Eq:Influence}. %The AFA baseline is the mean value found in the data set and for Shapley values it is the mean value of the studied output that forms the baseline for Shapley explanations. In classification tasks, the Shapley value baseline corresponds to the distribution of different classes in the training data. This difference regarding the baseline concept leads to differences in the $\phi$ values obtained using contextual influence versus AFA methods. 

\paragraph{Estimation of CI and CU}\label{Sec:EstimationCIU}

Most model-agnostic post-hoc XAI methods only attempt to estimate the importance/influence of one feature $i$ on the output, with the assumption that features are independent. In this paper, we limit the scope to that case and do not consider coalitions of features or CIU's \textit{intermediate concepts} as presented in \cite{FramlingThesis_1996,FramlingAISB_1996,Framling_EXTRAAMAS2020}\footnote{Intermediate concepts also deal with dependencies between features. However, in this paper we assume that the features are independent, as is the case for Shapley value and LIME too.}. Therefore, we only need to consider the case when $\{i\}$ has one single index and the case when $\{I\}=1,\dots,N$ in Equations~\ref{Eq:CI_definition}, \ref{Eq:CU} and \ref{Eq:CIU_InfluenceFinal} (and Equations~\ref{Eq:OriginalCI} and \ref{Eq:CU_y}).
$\{I\}=1,\dots,N$ signifies all inputs jointly, which by definition means that $umin_{j,\{I\}}=0$ and $umax_{j,\{I\}}=1$ for all instances $x$. Therefore, $umax_{j,\{I\}}(x)-umin_{j,\{I\}}(x)=1  \enspace \forall{x}$. If we have a classification task where $u_{i}(y_{j})=y_{j}$, then we also have $ymax_{j,\{I\}}(x)-ymin_{j,\{I\}}(x)=1  \enspace \forall{x}$ and in a regression task with $u_{j}(y_{j})=Ay_{j}+b$, $ymin_{j,\{I\}}$ and  $ymax_{j,\{I\}}$ are calculated accordingly. If the actual model gives values outside this range, then the model is over-shooting at least in some parts of the input space.  

\begin{figure*}
    \centering
    \begin{tabular}{cc}
        \includegraphics[width=0.45\textwidth]{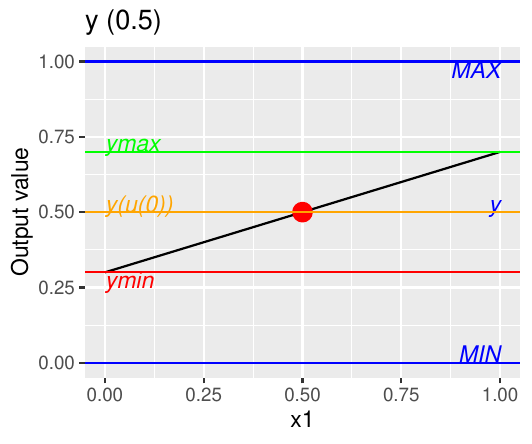}
        & 
        \includegraphics[width=0.45\textwidth]{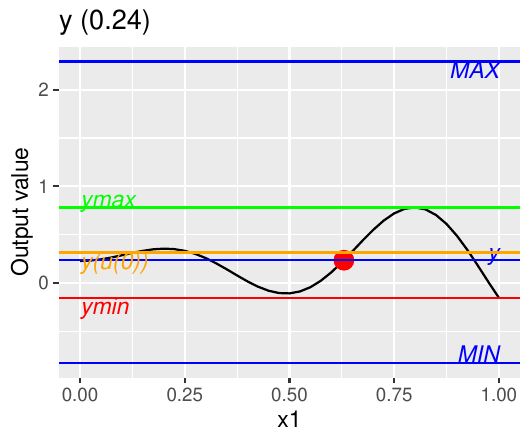} \\ 
        Linear, feature $x_{1}$ & Non-linear, feature $x_{1}$ \\ \includegraphics[width=0.45\textwidth]{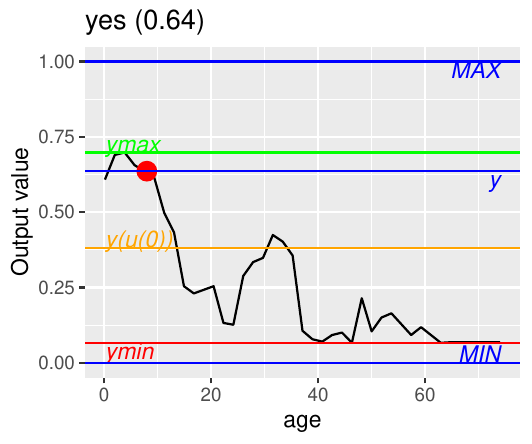} 
        &
        \includegraphics[width=0.45\textwidth]{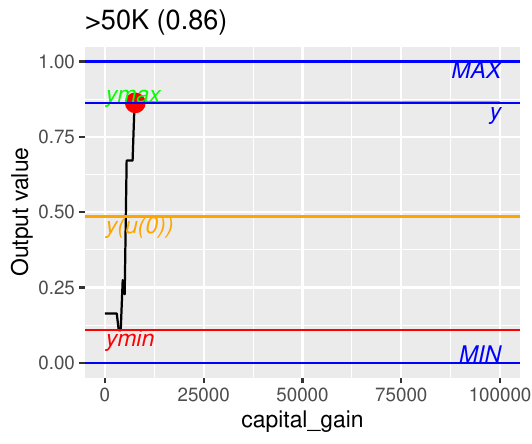} 
        \\
        Titanic, feature ``age'' & Adult, feature ``capital\_gain''
    \end{tabular}
    \caption{Output value $y$ as a function of one feature value for the four models, with illustration of CIU calculations. The red dot shows the $x_{i}$ value of the instance to be explained for the feature $i$. The output range $[MIN,MAX]$ is $[0,1]$ for all models, except for the ``non-linear'' model where it is $[-0.825,2.29]$. The labels in the Figure are $MIN=ymin_{j}$, $MAX=ymax_{j}$, $ymin=ymin_{j,\{i\}}(x)$, $ymax=ymax_{j,\{i\}}(x)$, $y=y_{j}(x)$ and $y(u(0))=ymin+\phi_{0}(ymax-ymin)$.} 
    \label{Fig:InputOutput_CIU}
\end{figure*}

%Calculating CI and CU values requires finding the minimal and maximal values  $umin_{j}$ and $umax_{j}$ . In practice, the output value utility function is almost always linear of the form $u_{j}(y_{j})=Ay_{j}+b$ or even $u_{j}(y_{j})=y_{j}$ for classification tasks. Therefore, we actually need to find the minimal and maximal output values $ymin_{j}$ and $ymax_{j}$ in . The fidelity and accuracy of CIU depends entirely on the correctness of these values.

The approach proposed in \cite{Framling_XAI_WS_AAAI_2021} is used for estimating $ymin$ and $ymax$\footnote{The approach in \cite{Framling_XAI_WS_AAAI_2021} is applicable to any feature set $\{i\}$ and $\{I\}$, including $1,\dots,N$.}. For categorical features, the approach uses all possible values. For numerical features the model is sampled using a set of instances consisting of \textbf{1)} the instance $x$, \textbf{2)} the instance $x$ with feature $i$ value replaced by the smallest possible value for feature $i$ ($min_{i}$), \textbf{3)} the instance $x$ with feature $i$ value replaced by the greatest possible value for feature $i$ ($max_{i}$), and \textbf{4)} a set of instances where the value of feature $i$ is replaced with a random value from the interval $[min_{\{i\}},max_{\{i\}}]$. This approach guarantees exact values for $ymin_{j}$ and $ymax_{j}$ if $f(x)$ is monotonous, or in practice if $ymin_{j}$ and $ymax_{j}$ values are found at the input values $min_{\{i\}}$ and $max_{\{i\}}$. If $min_{\{i\}}$ and $max_{\{i\}}$ values are not pre-defined, then they can be determined from the training data set. The calculation of CI and CU values with different data sets and models is illustrated in Fig.~\ref{Fig:InputOutput_CIU} using input/output value plots like in \cite{FramlingAISB_1996}, also called \textit{ceteris paribus} or \textit{what-if} plots in \cite{BiecekBurzykowski_Book_2021}. CIU values can be ``read'' directly from such plots, which makes CIU transparent at least when compared to AFA methods that might be considered black-boxes themselves. 

%The used approach for estimating $ymin_{j}$ and $ymax_{j}$ (and $umin_{j}$ and $umax_{j}$) will produce samples that might not be included in the training set of a ML model. That means that the model $f$ might not be ``competent'' to provide accurate values for those samples. Whether this is a 

The sampling approach that is used can lead to so called out of distribution (OOD) samples, \ie feature value combinations that are significantly different from the data in the training set used to build an ML model. For such samples, the model $f$ may be incapable to provide correct output values $y_{j}$. OOD challenges related to the used sampling method and potential solutions to those challenges can be grouped into at least the three following cases:

\begin{enumerate}
    \item \textbf{Predictable OOD behaviour.}  If OOD samples do not lead to undershooting of the $ymin$ value, nor to overshooting of the $ymax$ value, then OOD is not an issue. Ensemble learning models typically do not under- or overshoot even when extrapolating outside the training set. In \cite{FramlingAISB_1996} for instance, CIU was used with an radial basis function (RBF) net that guaranteed that under- or overshooting does not occur. Input-output value graphs such as those in Fig.~\ref{Fig:InputOutput_CIU} can be used for studying the model behaviour within the value ranges used by CIU. 
    \item \textbf{Non-predictable OOD behaviour.} This happens if under- or overshooting may occur with OOD samples. In that case the sampling approach used here will not be appropriate. Various approaches could be imagined for addressing this problem, such as only using samples that are "sufficiently" close to samples in the training set. 
    \item \textbf{Detecting model instability}. Since CI and CU values are in the range $[0,1]$ by definition, obtaining values that are outside this range indicate that the model undershoots or overshoots the permissible range for one or more samples. This could be an indication that those samples should be removed or that the model should be corrected in order to increase its trustworthiness. A correction approach using what what is called pseudo-examples was proposed in \cite{FramlingThesis_1996}. 
\end{enumerate}

\noindent The second and third cases are considered to be out of scope for the current paper and remain topics of further research. It is also worth mentioning that similar OOD challenges exist for all model-agnostic XAI methods (at least for the permutation-based ones), including Shapley value and LIME. We do not consider model-specific methods here, such as TreeSHAP for Shapley values \cite{LundbergEtAl_TreeSHAP_2018}. %We do, however, compare with the currently most used model-agnostic post-hoc XAI methods that are also permutation-based, \ie Shapley value and LIME.

\iffalse
\begin{algorithm}[t]
\SetAlgoLined
\LinesNumbered
    \KwResult{$N \times M$ matrix $S(x,\{i\})$}
    \Begin{
    \ForAll{categorical features}{
        $D \leftarrow$ all possible value combinations for categorical features $\{i\}$\;
        Randomize row order in $D$\;
        \If{$D$ has more rows than $N$}{Set $N$ to number of rows in $D$\;}
    }
    \ForAll{numerical features}{
        Initialize $N \times M$ matrix $R$ with input values $x$\;
        $R \leftarrow$ two rows per numerical features in $\{i\}$ where the current value is replaced by the values $min_{\{i\}}$ and $max_{\{i\}}$ respectively\;
        $R \leftarrow$ fill remaining rows to $N$ with random values from intervals $[min_{\{i\}},max_{\{i\}}]$;
    }
    $S(C,\{i\}) \leftarrow$ concatenation of $x$ with merged $D$ and $R$, where $D$ is repeated if needed to obtain $N$ rows\;
    }
    \caption{Constructing set of representative input vectors for estimating $ymin_{j}$ and $ymax_{j}$.}
    \label{algo:SRIV}
\end{algorithm}
\fi

\section{Results}\label{Sec:Experiments}

All used software is written in R and is available as open-source on Github, including the source code for producing the results shown here. The caret package~\cite{Caret_2008_Kuhn} is used for all machine learning models. %All models have been trained using 10-fold cross validation. 
The IML package is used for Shapley value calculations \cite{MolnarIML_2018} and the `lime' package for LIME \cite{LimeR}. The CIU implementation and results use the `ciu' package  \cite{Framling_XAI_WS_AAAI_2021} as a base. The default parameters are used for all packages unless stated otherwise. The experiments were run using Rstudio Version 1.3.1093 on a MacBook Pro, with 2.3 GHz 8-Core Intel Core i9 processor, 16 GB 2667 MHz DDR4 memory, and AMD Radeon Pro 5500M 4 GB  graphics card. 

\begin{figure*}
    \centering
    \begin{tabular}{lccc}
        \multirow{1}{*}[+6ex]{\rotatebox{90}{\smaller Linear}} & \includegraphics[width=0.31\textwidth]{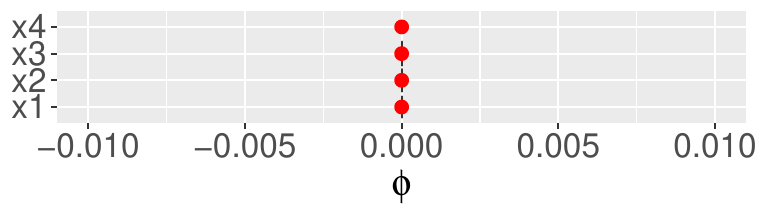}
        & \includegraphics[width=0.31\textwidth]{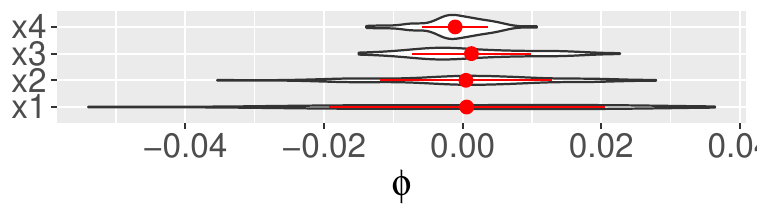} & \includegraphics[width=0.31\textwidth]{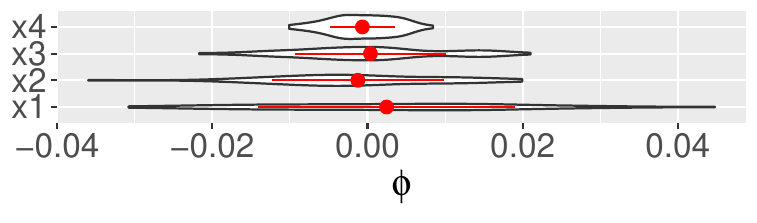} \\
        \multirow{1}{*}[+8ex]{\rotatebox{90}{\smaller Non-linear}} & \includegraphics[width=0.31\textwidth]{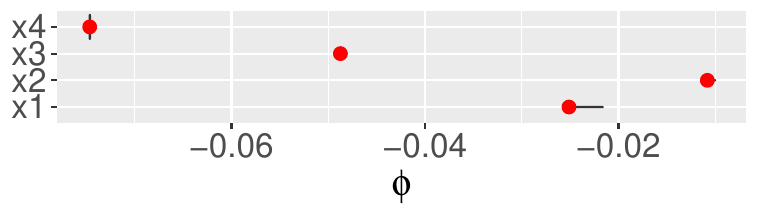}
         & \includegraphics[width=0.31\textwidth]{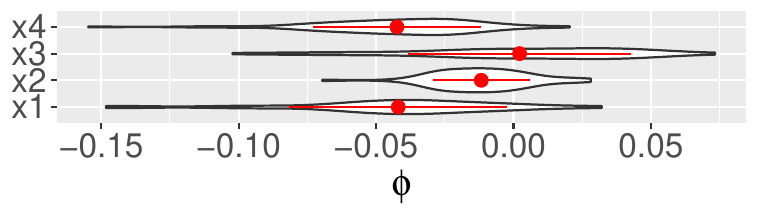} & \includegraphics[width=0.31\textwidth]{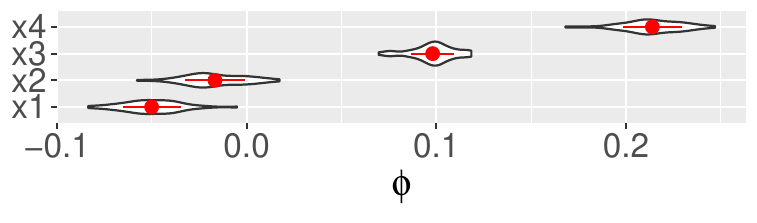} \\    
        \multirow{1}{*}[+8ex]{\rotatebox{90}{\smaller Titanic}} & \includegraphics[width=0.31\textwidth]{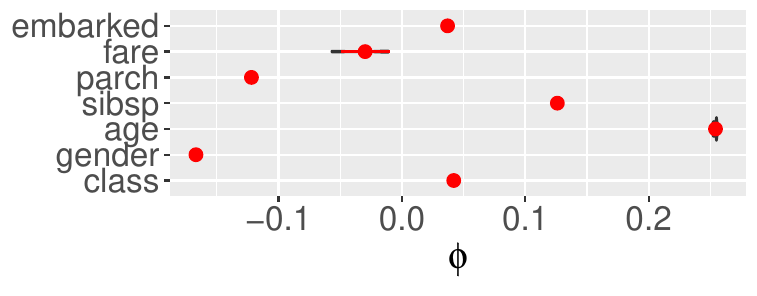}
         & \includegraphics[width=0.31\textwidth]{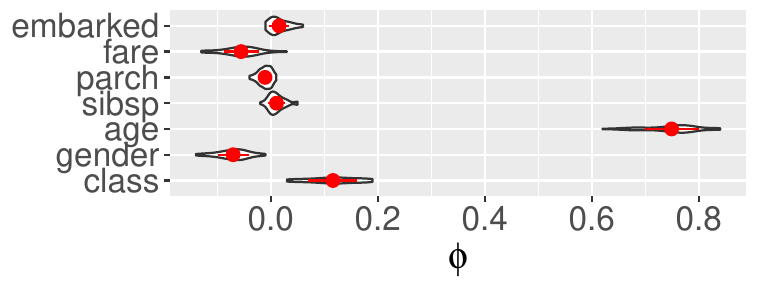} & \includegraphics[width=0.31\textwidth]{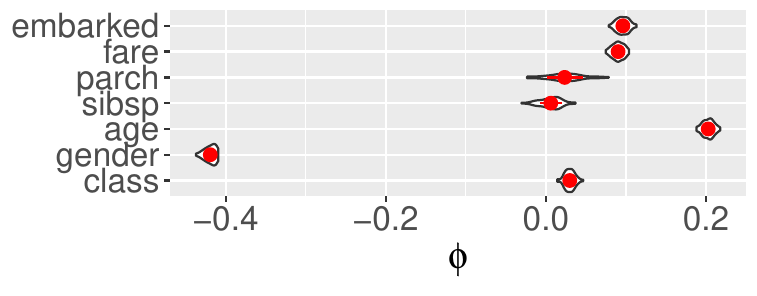} \\
         \multirow{1}{*}[+12ex]{\rotatebox{90}{\smaller Adult}} &
         \includegraphics[width=0.31\textwidth]{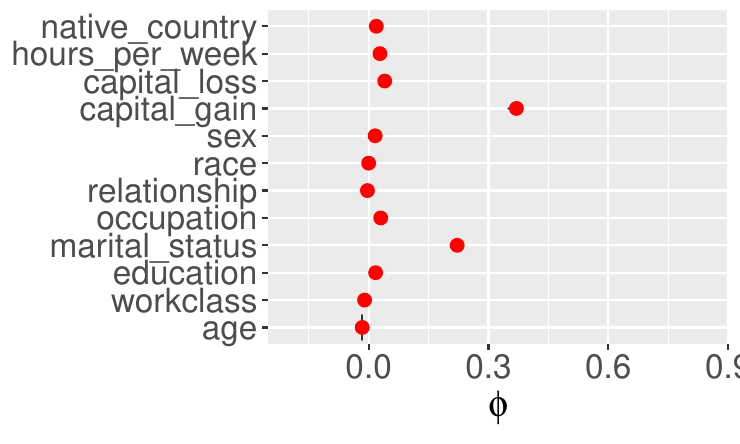} & \includegraphics[width=0.31\textwidth]{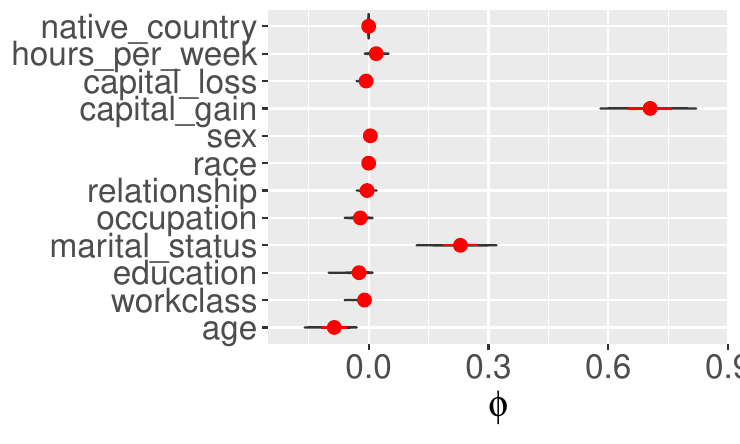} & \includegraphics[width=0.31\textwidth]{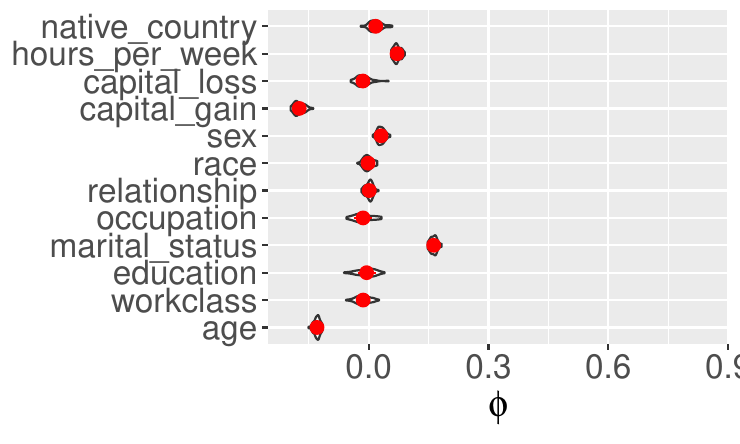} \\
         & Contextual influence & Shapley & LIME
    \end{tabular}
    \caption{Distribution of $\phi$ values from 50 runs with the studied data sets / models and the studied methods.}
    \label{fig:Sensitivity}
\end{figure*}

\begin{sloppypar*}
\noindent The ``Elapsed'' times shown in the results are indicated as ``real time'' (``elapsed'' value of $system.time()$ function) and have been collected for all methods during the same session in order to keep them as comparable as possible. Therefore, the exact values are not significant but only the ratio between the execution times of the different methods. It should also be emphasized that execution times do not depend only on the method itself but also on the used implementation and other factors. Parameters such as sample size have been set to be identical for all the methods. However, for sampling-based methods there is always a compromise between sample size (and execution time) versus the precision (variance) of the result. Therefore, the expectation is that the longer the execution time, the lower the variance. So a method with short execution time and low variance is preferred to a method with long execution time and high variance. 
\end{sloppypar*}

In order to make global feature importance values comparable, they have been normalised to the range $[0,1]$ for PFI and Shapley value by dividing with the sum of importance of all features. CI values are by definition in the range $[0,1]$. Unlike Shapley values, it doesn't seem like LIME values would have been used, or proposed to be used, for estimating global feature importance. Therefore LIME has not been included in the global feature importance experiments and results. 

For the instance-specific experiments, instances that have average (or close to  average) feature values have been chosen, except for the Titanic instance where we use the same instance as in \cite{BiecekBurzykowski_Book_2021}. The reason for this choice is that such instances illustrate the difference between importance and influence as clearly as possible. The importance $\omega_{i}$ of a feature does not depend on the feature's value $x_{i}$, whereas its influence value $\phi_{i}$ depends on $x_{i}$ in a way that gives close to zero $\phi_{i}$ values when $x_{i}$ is ``average''.

%Not using \textit{shapr}~\cite{Sellereite2019} package because ``Note that both the features and the prediction must be numeric. The approach is constructed for continuous features.'' 

\subsection{Known Linear Function}

We begin with a known linear function, for which we know the feature importances (weights) $w_{i}$ as well as the $E(X_{i})$ values. Therefore we know the correct results for CI, CU, Contextual influence and Shapley value. In addition to these, we include experiments with LIME. The function is:

\begin{equation*}
    f(x)=0.4x_{1} + 0.3x_{2} + 0.2x_{3} + 0.1x_{4}
\label{Eqn:ExpLinearFunction}
\end{equation*}

\begin{table}
\caption{Global importance of features for the known linear function, averaged over 50 iterations. CI and Shapley used 1000 randomly selected instances at each iteration. CI values are identical for all instances, so CI has zero variance.}
\label{Tab:GlobalLinearImportances}
\setlength{\tabcolsep}{8pt}
\begin{center}
\begin{tabular}{lcccc}
\toprule
Feature & PFI-MAE & 
%PFI-RMSE & 
CI & Shapley \\
\midrule
$x_{1}$ & $0.4\pm0.0$ & 
%$0.4\pm0.0$ & 
$0.4\pm0$ & $0.40\pm0.05$ \\
$x_{2}$ & $0.3\pm0.0$ & 
%$0.3\pm0.0$ & 
$0.3\pm0$ & $0.30\pm0.04$\\
$x_{3}$ & $0.2\pm0.0$ & 
%$0.2\pm0.0$ & 
$0.2\pm0$ & $0.20\pm0.04$ \\
$x_{4}$ & $0.1\pm0.0$ & 
%$0.1\pm0.0$ & 
$0.1\pm0$ & $0.10\pm0.02$ \\
\midrule
\textit{Elapsed} & \textit{1460s.} & 
%\textit{1588s.} & 
\textit{713s.} & \textit{27786s.} \\
\bottomrule
\end{tabular}
\end{center}
\end{table}

\noindent CIU can use the function directly as the studied model, whereas the other methods require the availability of a training set and a trained model. The training data set consisted of all possible value combinations of the four features $x_{i}$ in the range $[0,1]$ with a step of $0.05$, \ie 194481 instances. The trained linear model achieved $R^2=1$. All results are reported for the trained linear model. 

%Lin_FI_MAE_time50
%    user   system  elapsed 
%1144.234  314.164 1460.380 
%Lin_FI_MSE_time50
%    user   system  elapsed 
%1169.046  327.936 1501.071
%Lin_FI_RMSE_time50
%    user   system  elapsed 
%1236.176  347.164 1588.827
%Lin_CIU_time50
%   user  system elapsed 
%707.992   4.258 713.603
%Lin_Shap_time50
%    user   system  elapsed 
%26115.81  1643.61 27786.09

\paragraph{Global feature importance.} 

Table~\ref{Tab:GlobalLinearImportances} shows results for the PFI method with the Mean Average Error (MAE) loss function, $mean(CI)$ and $mean|Shapley~value|$. All methods retrieve the original weights of the linear model but with different accuracy and variance. CI values are identical for all instances in the case of linear models. Therefore, \textit{importance as defined by CI is conceptually identical with global feature importance}. Even though Shapley values estimate instance-level influence it still gives similar values as the other methods but with a high variance, despite a significantly longer execution time.

\paragraph{Instance-specific results.}

\begin{table}[t]
\caption{Local importance, utility and influence values by different methods for the known linear function. The numbers show mean values over 50 iterations.}
\label{Tab:LinearImportances}
\setlength{\tabcolsep}{8pt}
\begin{center}
\begin{tabular}{lccccc}
\toprule
Feature & CI & CU & $\phi_{CIU}$ & $\phi_{Shapley}$ & $\phi_{LIME}$ \\
\midrule
$x_{1}=0.5$ & 0.4 & 0.5 & 0 & ~0.004 & -0.003 \\
$x_{2}=0.5$ & 0.3 & 0.5 & 0 & -0.001 & ~0.001 \\
$x_{3}=0.5$ & 0.2 & 0.5 & 0 & -0.001 & ~0.001 \\
$x_{4}=0.5$ & 0.1 & 0.5 & 0 & ~0.001 & -0.002 \\
\bottomrule
\end{tabular}
\end{center}
\end{table}

\begin{figure*}
    \centering
    \setlength{\tabcolsep}{1pt}
    \begin{tabular}{cccc}
        \vspace{-5pt}
        \includegraphics[width=0.35\textwidth]{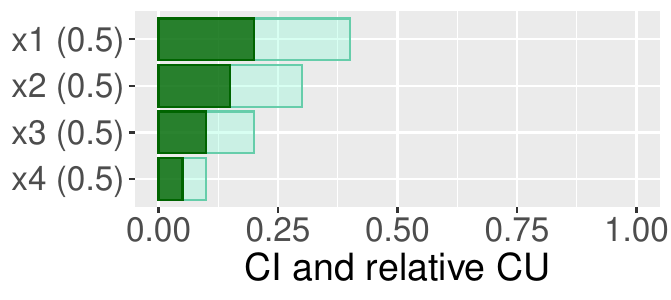}
        & \includegraphics[width=0.35\textwidth]{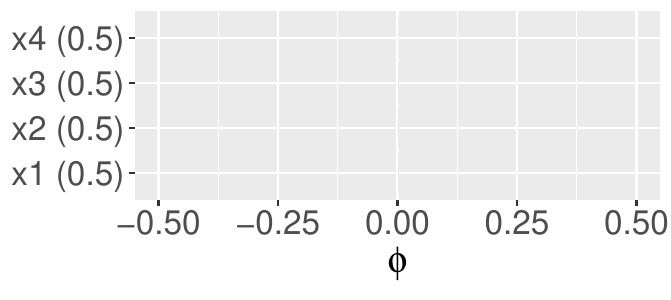} \\ 
        \vspace{5pt}
        CIU & Contextual influence  \\ 
        \vspace{-5pt}
        \includegraphics[width=0.35\textwidth]{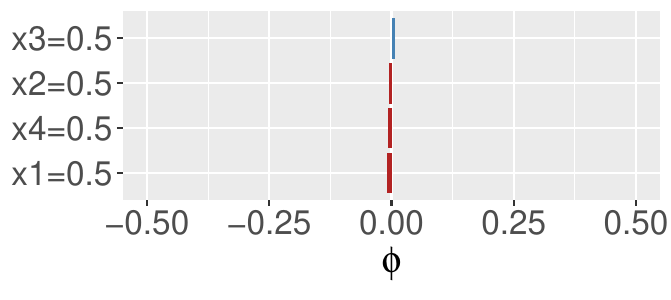} 
        &
        \includegraphics[width=0.35\textwidth]{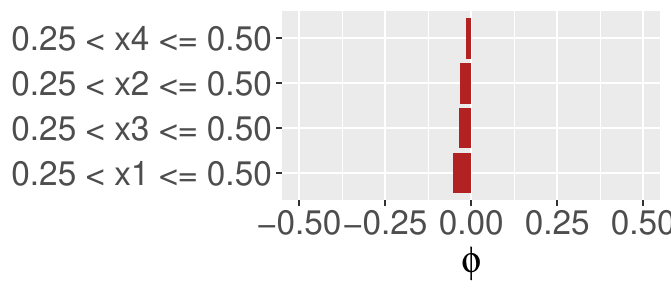} \\
        Shapley value & LIME
    \end{tabular}
    \caption{Bar plot explanations for linear model and studied instance with different methods. For CIU and Contextual influence the plot is identical on consecutive runs. For Shapley value and LIME the plot changes on every run due to their variance.}
    \label{Fig:LinearExplanations}
\end{figure*}

We study the instance $x=0.5,0.5,0.5,0.5$ with $f(x)=0.5$ for which $\phi_{i}$ values are zero by definition. Table~\ref{Tab:LinearImportances} shows the CI, CU and $\phi$ values obtained. %that CI values are the same for all instances and consistently indicate the potential effect of changing the value of each feature. %For an eliminated applicant, the best choice for improvement would presumably be input features with a high CI value and a low CU value, \ie input features that could improve the outcome value the most.  
%On the other hand, an influence-based explanation would in this case be a ``null explanation'', which says that ``this instance is average because all input features are average''. Such a null explanation is insufficient in many situations because it filters out the potential effect of modifying the value of an average input feature. %For instance, providing such an explanation to an eliminated applicant for a work position would in no case suggest improving average-value features even though they might be the most valuable to focus on in order to improve the chances in the future. 
Fig.~\ref{fig:Sensitivity} shows the same values and their variance over 50 runs for all the methods. CIU/Contextual influence has exactly zero variance. Shapley values and LIME have a great variance, so for the studied instance even the sign of $\phi_{i}$ values will change randomly from one explanation to the next. This signifies that the Shapley value and LIME plots in Fig.~\ref{Fig:LinearExplanations} are misleading. Furthermore, the influence-based explanations don't provide hardly any information, whereas the CIU plot does provide useful information about the model and the result. 

The CIU plot in Fig.~\ref{Fig:LinearExplanations} illustrates how CI and CU can be combined into a more information-rich explanation than with influence alone. The length of the half-transparent bar corresponds to the CI value and shows how much modifying the value of the feature would modify the output. The solid bar illustrates the CU value so that a $CU=1$ value will cover the transparent bar entirely, while $CU=0$ gives a zero-length solid bar. Therefore, the solid bar shows ``how good'' the current value is compared with the worst and best possible values for the feature. CI and CU values can be visualised in many ways but the current one has been selected for its ``counterfactual'' aspect, which signifies that it indicates which features would have the greatest potential for improving the output utility. Such functionality is useful for instance if getting a negative credit decision and seeing what criteria could have the greatest effect if it would be possible to improve the values of those criteria. 

\subsection{Known Non-linear Function}

% \begin{equation*}
%     \resizebox{\linewidth}{!}{$
%     \displaystyle
%     f(x)=0.7x_{1}sin(10x_{1}) + 0.3x_{2}sin(10x_{2}) + x_{3}^2 + 
%     (2x_{4}^4 - 1.5x_{4}^2)$}
% \label{Eqn:NonlinearFunction}
% \end{equation*}

In order to study the behaviour with a known, non-linear function we use the function in Equation~\ref{Eqn:NonlinearFunction}. It is worth noting that all features are independent in this function. A Stochastic Gradient Boosting model was trained with a similar training data as for the linear function, \ie 194481 instances and achieved $R^2=0.992$. 

\begin{align}
    f(x)=0.7x_{1}sin(10x_{1}) + 0.3x_{2}sin(10x_{2}) + 
%    \nonumber\\
    x_{3}^2 + (2x_{4}^4 - 1.5x_{4}^2)
\label{Eqn:NonlinearFunction}
\end{align}

%min(y): -0.8248211
%max(y): 2.291487

%Stochastic Gradient Boosting (caret `gbm' model):      RMSE   Rsquared        MAE 
%0.04443707 0.99158403 0.03437012
%$f(x)=0.235$ for the studied input. 

\begin{table}
\caption{Global importance of input features in percent for the known non-linear function, averaged over 50 iterations. CI and Shapley used 1000 randomly selected instances at each iteration.}
\label{Tab:GlobalNonlinearImportances}
\setlength{\tabcolsep}{8pt}
\begin{center}
\begin{tabular}{lcccc}
\toprule
Feature & PFI-MAE & 
%PFI-MSE & 
CI & Shapley \\
\midrule
$x_{1}$ & $30.5\pm0.0$ & 
%$31.6\pm0.0$ & 
$29.8\pm0.1$ & $29.4\pm0.6$ \\
$x_{2}$ & $13.0\pm0.0$ & 
%$5.7\pm0.0$ & 
$11.7\pm0.0$ & $11.6\pm$ 0.3\\
$x_{3}$ & $36.4\pm0.0$ & 
%$46.1\pm0.0$ & 
$33.3\pm0.1$ & $39.5\pm0.6$ \\
$x_{4}$ & $20.1\pm0.0$ & 
%$16.6\pm0.0$ & 
$25.2\pm0.0$ & $19.6\pm0.5$ \\
\midrule
\textit{Elapsed} & \textit{1861s.} & 
%\textit{1951s.} & 
\textit{618s.} & \textit{35501s.} \\
\bottomrule
\end{tabular}
\end{center}
\end{table}

\paragraph{Global feature importance.} 

Based on the results in Table~\ref{Tab:GlobalNonlinearImportances} it seems like all methods agree on the order of importance but it is not possible to say which one is the most ``correct'' one because they are all based on slightly different definitions of what global feature importance signifies. In \cite{LundbergEtAl_TreeSHAP_2018} it is claimed that using average importance of all instances provides a better estimate of the global importance than PFI, for instance. That paper uses mean absolute Shapley values. However, as shown in this paper, CI provides a ``true'' importance measure, rather than the influence values given by Shapley values. CI is orders of magnitude faster than Shapley values and still gives lower variance.

\paragraph{Instance-specific results.}

\begin{sloppypar*}
We choose to study the instance $x=0.63,0.63,0.59,0.81$ with $f(x)=0.235$, which is close to the average $f(x)$ value and therefore gives low expected $\phi_{i}$ values. 
%Table~\ref{Tab:NonlinearImportances} shows that instance-level CI values are almost identical with the CI-based global feature importance in Table~\ref{Tab:GlobalNonlinearImportances}, which again empirically indicates that CI is indeed conceptually identical on the instance level and on the global level. %Even though CI may also be instance-specific, it is CU that indicates to what extent the instance's feature values contribute to a high output utility value. Finally, as shown by 
Fig.~\ref{fig:Sensitivity} shows that Contextual influence has close to zero variance, which is therefore true also for CI and CU. Influence-based methods all give close to zero $\phi_{i}$ values, resulting in explanations with a low information value, as shown in Fig.~\ref{Fig:NonLinearExplanations}. The great variance of Shapley values may again cause the signs of $\phi_{i}$ to change from one run to the other, which is true also for LIME. %LIME's variance is smaller but the mean $\phi_{i}$ values from the Shapley values. 
\end{sloppypar*}

\begin{table}
\caption{Local importance, utility and influence values given by different methods for the known non-linear function, averaged over 50 iterations.}
\label{Tab:NonlinearImportances}
\setlength{\tabcolsep}{8pt}
\begin{center}
\begin{tabular}{lccccc}
\toprule
Feature & CI & CU & $\phi_{CIU}$ & $\phi_{Shapley}$ & $\phi_{LIME}$ \\
\midrule
$x_{1}=0.63$ & 0.300 & 0.416 & -0.025 & -0.031 & -0.049 \\
$x_{2}=0.63$ & 0.128 & 0.416 & -0.011 & -0.015 & -0.021 \\
$x_{3}=0.59$ & 0.321 & 0.348 & -0.049 & ~0.011 & ~0.095 \\
$x_{4}=0.81$ & 0.251 & 0.202 & -0.075 & -0.040 & ~0.216 \\
\bottomrule
\end{tabular}
\end{center}
\end{table}

\begin{figure*}
    \centering
    \setlength{\tabcolsep}{1pt}
    \begin{tabular}{cccc}
        \vspace{-5pt}
        \includegraphics[width=0.35\textwidth]{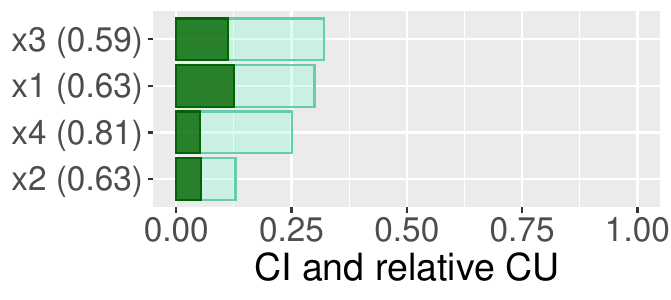}
        & \includegraphics[width=0.35\textwidth]{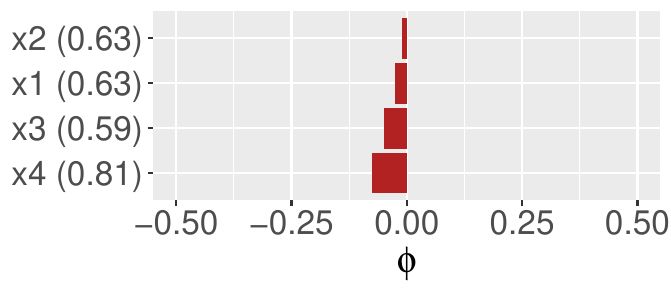} \\ 
        \vspace{5pt}
        CIU & Contextual influence  \\ 
        \vspace{-5pt}
        \includegraphics[width=0.35\textwidth]{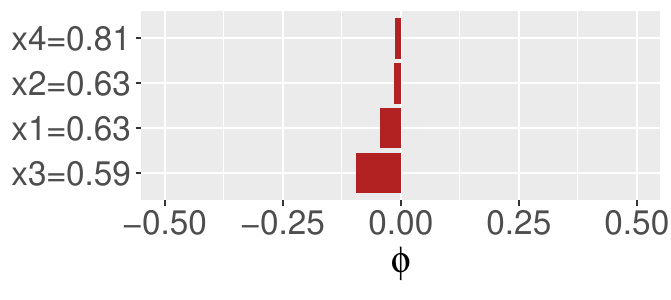} 
        &
        \includegraphics[width=0.35\textwidth]{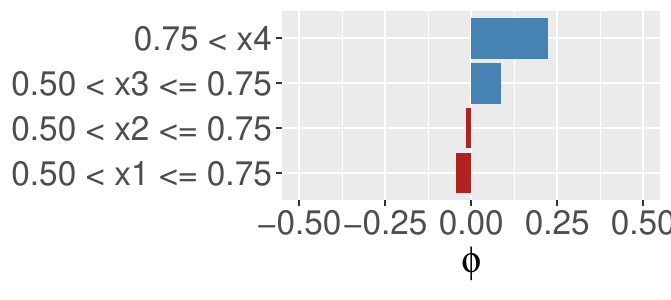} \\
        Shapley value & LIME
    \end{tabular}
    \caption{Bar plot explanations for non-linear model and studied instance with different methods. For CIU and Contextual influence the plot is identical on consecutive runs. For Shapley value and LIME the plot changes on every run due to their variance.}
    \label{Fig:NonLinearExplanations}
\end{figure*}

\subsection{Titanic}

\begin{figure*}
    \centering
    \setlength{\tabcolsep}{1pt}
    \begin{tabular}{cccc}
        \vspace{-5pt}
        \includegraphics[width=0.49\textwidth]{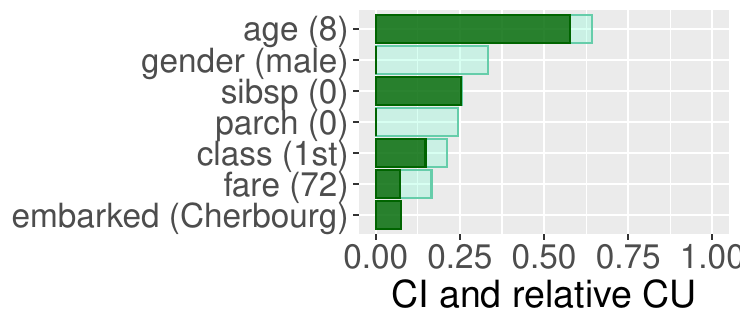}
        & \includegraphics[width=0.49\textwidth]{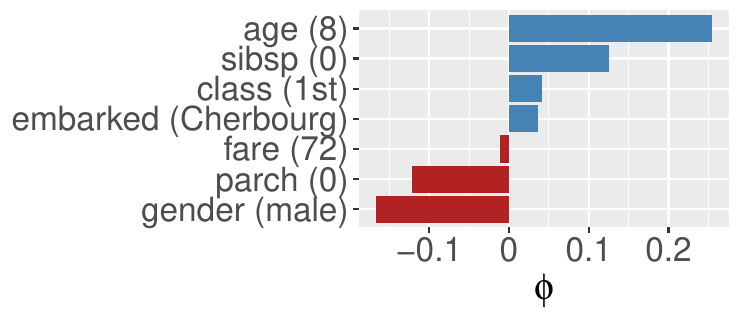} \\ 
        \vspace{5pt}
        CIU & Contextual influence  \\ 
        \vspace{-5pt}
        \includegraphics[width=0.49\textwidth]{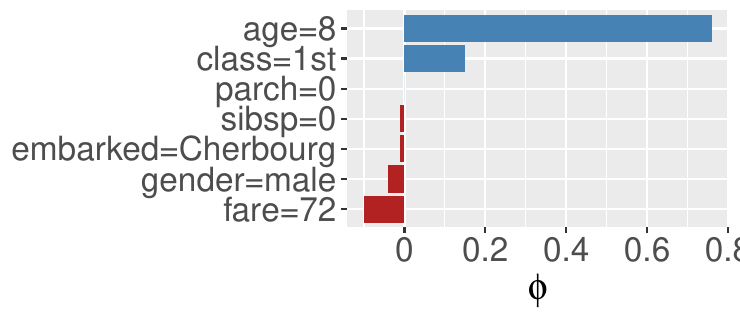} 
        &
        \includegraphics[width=0.49\textwidth]{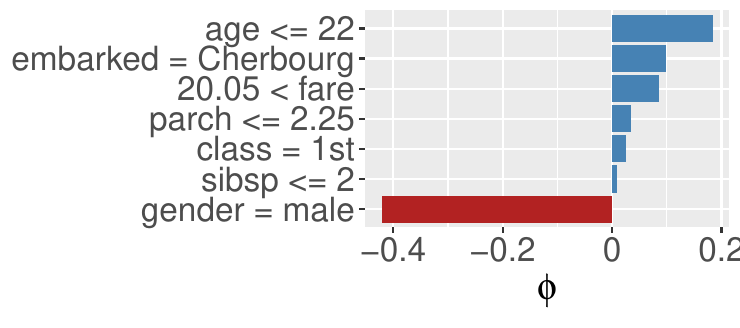} \\
        Shapley value & LIME
    \end{tabular}
    \caption{Bar plot explanations of 63.6\% survival probability of 8-year old boy `Johnny D' on Titanic with different methods.}
    \label{Fig:TitanicExplanations}
\end{figure*}

% Titanic_FIs_time50
%   user  system elapsed 
%187.664  15.067 203.288
% Titanic_CIU_time10
%    user   system  elapsed 
% 980.138  310.029 1292.762
% Titanic_Shap_time5
%    user   system  elapsed 
%1464.335   33.278 1500.986
% Column names: "class"    "gender"   "age"      "sibsp"    "parch"    "fare"     "embarked" "survived"
% CI standard deviations: 0.003764323 0.004634833 0.002625602 0.001784573 0.00114 0.002275463 0.002111757
% Shapley standard deviations: 0.012942573 0.012544104 0.009512269 0.003490852 0.002305949 0.006757460 0.003341111

The Titanic data set is a classification task with classes `yes' or `no' for the probability of survival. The data set has 2179 instances. We used a Random Forest model that achieved 81.1\% classification accuracy on the test set, which was 25\% of the whole data set. 

\paragraph{Global feature importance.} 

\begin{table}
\caption{Global importance of input features for Titanic data set, averaged over 50 iterations. CI and Shapley used 500 randomly selected instances at each iteration.}
\label{Tab:TitanicGlobal}
\setlength{\tabcolsep}{12pt}
\begin{center}
\begin{tabular}{lccc}
\toprule
Feature & PFI-CE & CI & Shapley \\
\midrule
Gender & $0.244\pm0.003$ & $0.236\pm0.005$ & $0.479\pm0.013$ \\
Class & $0.163\pm0.002$ & $0.254\pm0.004$ & $0.185\pm0.013$ \\
Age & $0.157\pm0.002$ & $0.227\pm0.003$ & $0.142\pm0.010$ \\ 
Fare & $0.152\pm0.002$ & $0.156\pm0.001$ & $0.107\pm0.007$ \\   
Embarked & $0.098\pm0.001$ & $0.059\pm0.001$ & $0.038\pm0.003$ \\
Sibsp & $0.096\pm0.000$ & $0.038\pm0.002$ & $0.033\pm0.003$ \\
Parch & $0.089\pm0.000$ & $0.029\pm0.001$ & $0.017\pm0.002$ \\
\midrule
\textit{Elapsed} & \textit{203s.} & \textit{6463s.} & \textit{15009s.}\\
\bottomrule
\end{tabular}
\end{center}
\end{table}

The class error (CE) loss function was used for PFI. Table~\ref{Tab:TitanicGlobal} shows that all methods agree on the order of feature importance, even though the least important features get lower importance estimates with CI and Shapley than with the global importance method. The importance value 0.479 given by Shapley values to the `gender' feature seems surprisingly high compared to the other methods. %The variance of CI is only slightly higher than for the global importance (CE) method but is much slower. 
CI again has significantly lower variance than Shapley values and is  faster.

%Shapley: Predicted value: 1.000000, Average prediction: 0.256269. Apparently at least the IML package uses 0/1 as predicted value for classification tasks! 

\begin{table}
\caption{Local importance/influence/utility estimations for Titanic instance `Johnny D', averaged over 50 iterations.}
\label{Tab:TitanicLocal}
\setlength{\tabcolsep}{8pt}
\begin{center}
\begin{tabular}{lccccc}
\toprule
Feature & CI & CU & $\phi_{CIU}$ & $\phi_{Shapley}$ & $\phi_{LIME}$ \\
\midrule
Gender & 0.334 & 0 & -0.334 & -0.070 & -0.419\\
Age & 0.637 & 0.899 & ~0.508 & ~0.749 & ~0.203 \\ 
Class & 0.212 & 0.698 & ~0.084 & ~0.116 & ~0.030 \\
Fare & 0.210 & 0.373 & -0.060 & -0.056 & ~0.090 \\   
Embarked & 0.074 & 1 & ~0.074 & ~0.015 & ~0.096 \\
Sibsp & 0.256 & 0.992 & ~0.252 & ~0.011 & ~0.006 \\
Parch & 0.244 & 0 & -0.244 & -0.011 & ~0.023 \\
\bottomrule
\end{tabular}
\end{center}
\end{table}

\paragraph{Instance-specific results.}

The studied instance is ``Johnny D'', an 8-year old boy from the Titanic that is also used and analyzed in \cite{BiecekBurzykowski_Book_2021}. Feature values and explanations by the different methods are shown as bar plots in Fig.~\ref{Fig:TitanicExplanations} for the probability of survival, which is 63.6\%. For Johnny D, the feature ``age'' has a clearly higher CI value than the global CI, which is normal because in the context of an 8-year old child the age is the most important feature, as shown by the input-output graph in Fig.~\ref{Fig:InputOutput_CIU}. Fig.~\ref{fig:Sensitivity} again shows that Contextual influence has close to zero variance. Shapley values and LIME again show great variance and seem to over-emphasize ``age'' (Shapley) and ``gender'' (LIME). 

%Only `age' and `fare' have non-zero standard deviation for Contextual influence (0.003428571 and 0.039763300). 

%titanic_sens_shap <- sens_analysis(new_passenger[,-ncol(titanic.train)],predictor,output.name = out.name)
%   class   gender      age    sibsp    parch     fare embarked 
%  0.1162  -0.0704   0.7492   0.0106  -0.0108  -0.0556   0.0154 
%     class     gender        age      sibsp      parch       fare   embarked 
%0.04579858 0.02927421 0.05033926 0.01517382 0.01209486 0.03264903 0.01820630
%titanic_sens_lime <- sens_analysis(new_passenger[,-ncol(titanic.train)],explainer,output.name = out.name)
%      class      gender         age       sibsp       parch        fare    embarked 
% 0.02979577 -0.41919916  0.20273462  0.00611742  0.02345552  0.09019055  0.09610838 
%      class      gender         age       sibsp       parch        fare    embarked 
%0.006347333 0.007038210 0.006967336 0.013758162 0.022676587 0.007004765 0.007661051

\subsection{Adult data set}

The Adult data set classifies people into the classes of yearly income in US dollars of ``$<=50K$'' and ``$>50K$''. A Stochastic Gradient Boosting model achieved 86.2\% classification accuracy on the test set. The test set contained 25\% of the whole data set. This data set is mainly included in order to validate the results with another ``real-life'' data set that has a greater number of features than Titanic. The Adult data set has 30162 instances, which is an order of magnitude more than for Titanic. 

% \begin{table}
% \centering
% \begin{tabular}{lccc}
% \toprule
% Feature & FI-CE & CI & CU \\
% \midrule
% capital\_gain & 0.102 & 0.370 & 0.000 \\
% marital\_status & 0.101 & 0.111 & 0.201 \\ 
% education & 0.089 & 0.097 & 0.274 \\ 
% occupation & 0.084 & 0.055 & 0.460 \\
% age & 0.084 & 0.055 & 0.608 \\
% capital\_loss & 0.082 & 0.171 & 0.023 \\
% hours\_per\_week & 0.078 & 0.051 & 0.381 \\
% relationship & 0.077 & 0.029 & 0.100 \\ 
% workclass & 0.076 & 0.013 & 0.415 \\
% sex & 0.076 & 0.013 & 0.600 \\
% native\_country & 0.076 & 0.033 & 0.360 \\
% race & 0.076 & 0.001 & 0.750 \\
% \bottomrule
% \end{tabular}
% \caption{Global importance estimations for Adult data set. All values are mean values over 50 iterations.}
% \label{Tab:AdultGlobalImportances}
% \end{table}

\begin{table}
\caption{Global importance of input features in percent for Adult data set, averaged over 50 iterations. CI and Shapley used 500 randomly selected instances at each iteration.}
\label{Tab:AdultGlobalImportances}
\setlength{\tabcolsep}{8pt}
\begin{center}
\begin{tabular}{lccc}
\toprule
Feature & PFI-CE & CI & Shapley \\
\midrule
marital\_status & $10.0\pm0.0$ & $10.4\pm0.2$ & $22.7\pm0.7$ \\ 
capital\_gain & $9.9\pm0.0$ & $29.9\pm0.7$ & $14.0\pm1.2$ \\
education & $8.9\pm0.0$ & $12.0\pm0.2$ & $17.8\pm0.7$ \\ 
age & $8.5\pm0.0$ & $7.8\pm0.2$ & $12.2\pm0.5$ \\
occupation & $8.3\pm0.0$ & $7.3\pm0.2$ & $12.8\pm0.4$ \\
capital\_loss & $8.2\pm0.0$ & $17.2\pm0.2$ & $5.0\pm0.5$ \\
hours\_per\_week & $7.8\pm0.0$ & $5.8\pm0.1$ & $7.7\pm0.3$ \\
relationship & $7.7\pm0.0$ & $3.3\pm0.1$ & $2.4\pm0.2$ \\ 
workclass & $7.7\pm0.0$ & $2.3\pm0.1$ & $2.2\pm0.2$ \\
sex & $7.6\pm0.0$ & $1.5\pm0.0$ & $2.7\pm0.2$ \\
native\_country & $7.6\pm0.0$ & $2.5\pm0.1$ & $0.5\pm0.1$ \\
race & $7.6\pm0.0$ & $0.0\pm0.0$ & $0.0\pm0.0$ \\
\midrule
\textit{Elapsed} & \textit{1291s.} & \textit{3314s.} & \textit{25111s.}\\
\bottomrule
\end{tabular}
\end{center}
\end{table}

\paragraph{Global feature importance.} 

Table~\ref{Tab:AdultGlobalImportances} shows similar results as for Titanic. CI gives a much higher importance to ``capital gain'' and ``capital loss'' features, which can be understood when studying the input-output graphs for those features and realizing that good values for either one of those features greatly increases the probability of the class ``$>50K$'' (see Fig.~\ref{Fig:InputOutput_CIU} for ``capital\_gain'' of the studied instance). Shapley values give a high importance to ``marital status'' that is difficult to explain. 

\paragraph{Instance-specific results.}

\begin{table}
\caption{Local importance/influence/utility estimations for Adult instance, averaged over 50 iterations.}
\label{Tab:AdultLocal}
\setlength{\tabcolsep}{8pt}
\begin{center}
\begin{tabular}{lccccc}
\toprule
Feature & CI & CU & $\phi_{CIU}$ & $\phi_{Shapley}$ & $\phi_{LIME}$ \\
\midrule
age & 0.384 & 0.944 & ~0.341 & ~0.117 & ~0.021 \\
workclass & 0.090 & 0.774 & ~0.049 & ~0.026 & ~0.005 \\
education & 0.433 & 0.987 & ~0.421 & ~0.224 & ~0.084 \\
marital\_stat. & 0.520 & 1.000 & ~0.520 & ~0.272 & ~0.164 \\
occupation & 0.320 & 1.000 & ~0.320 & ~0.125 & ~0.049 \\
relationship & 0.157 & 1.000 & ~0.157 & ~0.107 & ~0.058 \\
race & 0.000 & 0.000 & ~0.000 & ~0.000 & -0.001 \\
sex & 0.028 & 0.000 & -0.028 & -0.029 & -0.030 \\
capital\_gain & 0.241 & 0.218 & -0.118 & -0.012 & -0.180 \\
capital\_loss & 0.241 & 0.302 & -0.088 & -0.004 & -0.014 \\
hours\_per\_w. & 0.160 & 0.632 & ~0.042 & ~0.000 & -0.075 \\
native\_count. & 0.060 & 0.794 & ~0.035 & ~0.003 & ~0.018 \\
\bottomrule
\end{tabular}
\end{center}
\end{table}

%predict(gbm.adult,instance,type="prob")
%      <=50K      >50K
%1 0.1883055 0.8116945

\begin{figure*}
    \centering
    \setlength{\tabcolsep}{1pt}
    \begin{tabular}{cccc}
        \vspace{-5pt}
        \includegraphics[width=0.49\textwidth]{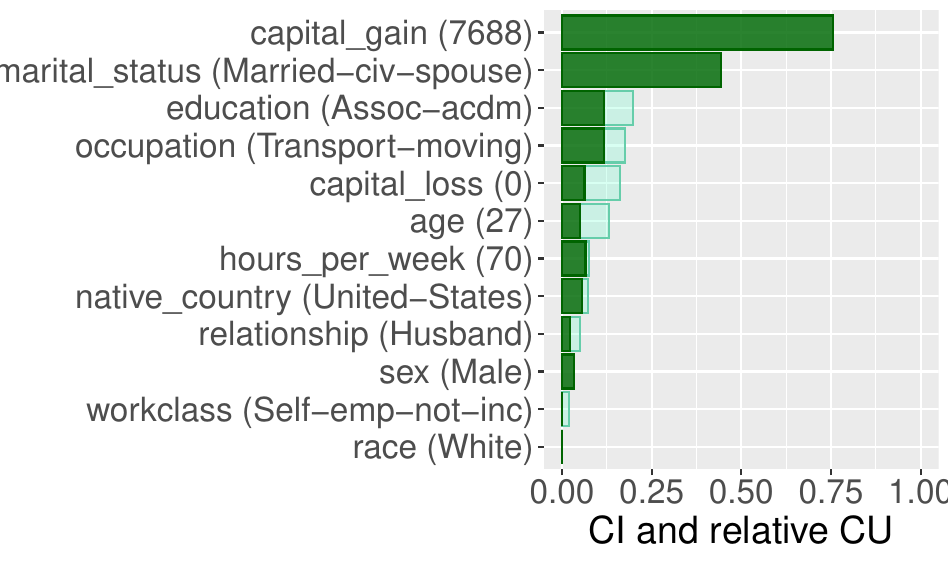}
        & \includegraphics[width=0.49\textwidth]{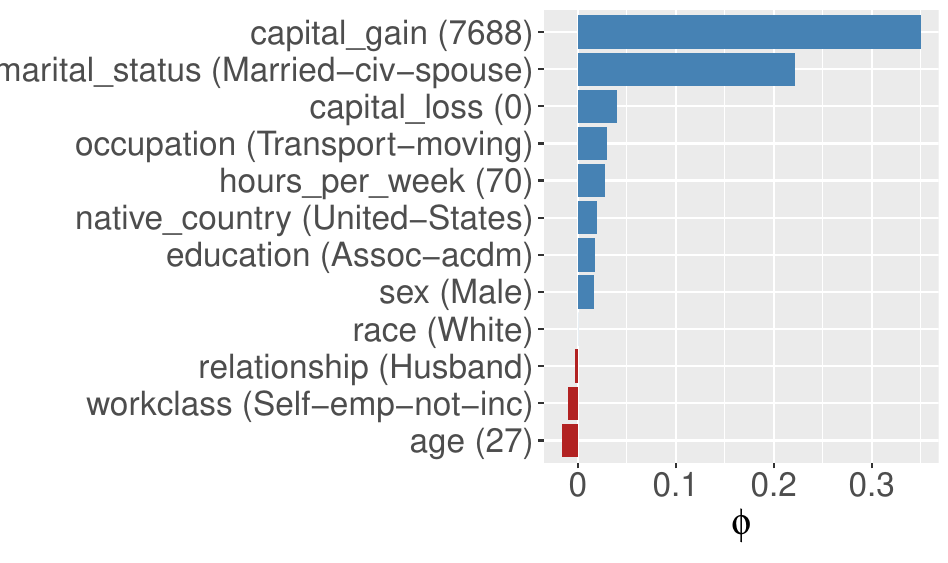} \\ 
        \vspace{5pt}
        CIU & Contextual influence  \\ 
        \vspace{-5pt}
        \includegraphics[width=0.49\textwidth]{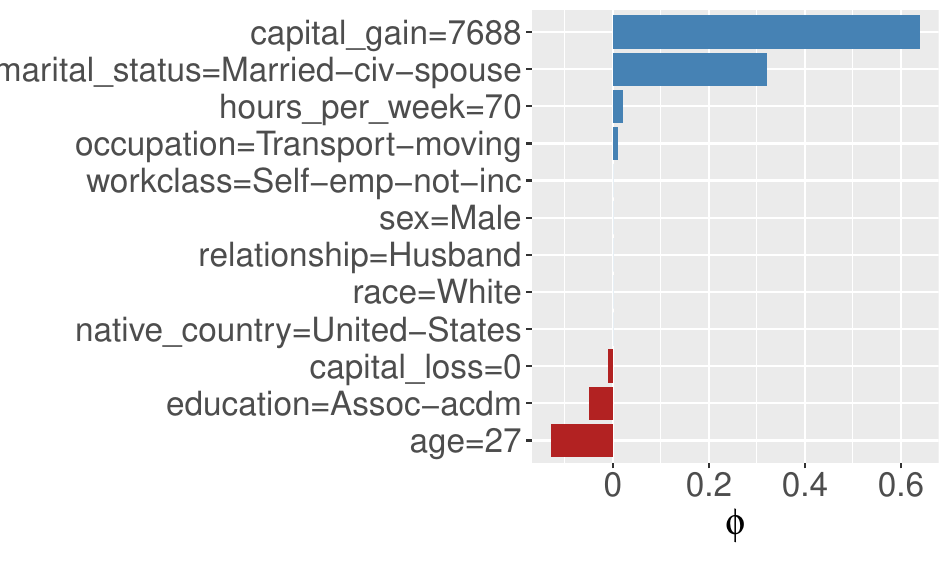} 
        &
        \includegraphics[width=0.49\textwidth]{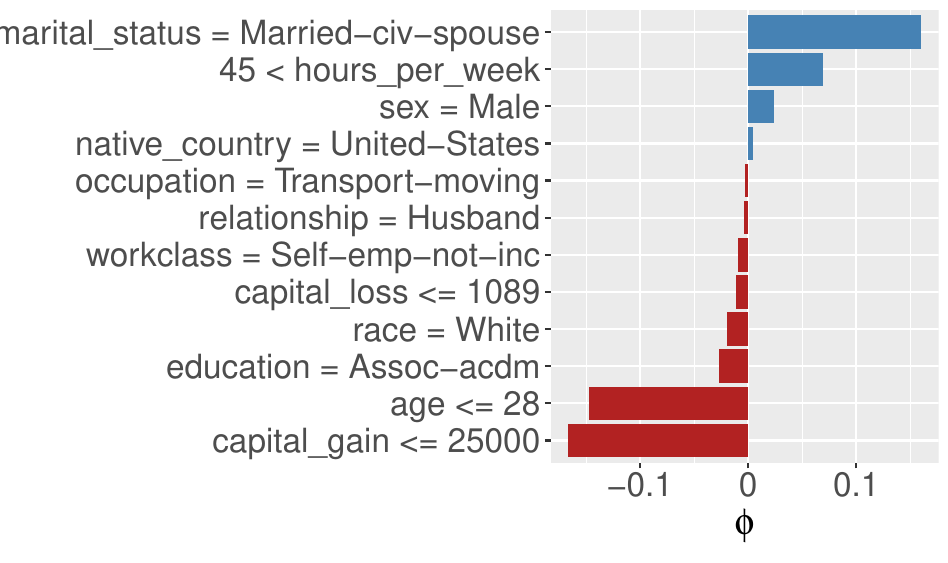} \\
        Shapley value & LIME
    \end{tabular}
    \caption{Bar plot explanations of 86.5\% probability of belonging to the class ``$>50K$'' with different methods.}
    \label{Fig:AdultExplanations}
\end{figure*}

The studied instance has a 86.3\% probability of belonging to the class ``$>50K$''. This particular instance has been chosen because the value ``$age=27$'' is unusual for a person that belongs to the class ``$>50K$'' and where the value for feature ``capital\_gain'' is  among the best possible and therefore makes this feature the most important/influential one. The high importance/influence of ``capital\_gain'' is shown by CIU, Contextual influence and Shapley value in Fig.~\ref{Fig:AdultExplanations}, whereas LIME shows radically different results. The Shapley values of many features are very close to zero, so it seems like the most influential features are over-emphasized in the same way as the ``age'' feature for the Titanic data set. 
Fig.~\ref{fig:Sensitivity} again confirms the differences in variance between the methods. 
%shows that Contextual influence has zero variance, except for `capital\_gain'. Shapley values and LIME again show great variance and difference in results. 
%Table~\ref{Tab:AdultLocal} shows that the CI value for ``age'' is much higher than what is given by Shapley values and LIME, which makes perfectly sense when studying the corresponding input-output graph in Fig.~\ref{Fig:InputOutput_CIU}.

\section{Conclusions}

As illustrated by the barplot ``explanations'' in Fig.~\ref{Fig:LinearExplanations}, Fig.~\ref{Fig:NonLinearExplanations}, Fig.~\ref{Fig:TitanicExplanations} and Fig.~\ref{Fig:AdultExplanations} CI and CU values can provide a kind of counterfactual ``what-if?'' explanations that show what features have the greatest potential to change the outcome, while also providing an indication of what features have values that could be improved. Such information is missing in the influence-based explanations, which only indicate positive or negative influence compared with a reference value. 

CI, CU and Contextual influence have known ranges and can therefore be interpreted directly. $CI=1$ signifies that the output can be modified completely by modifying the feature value and $CI=0$ signifies ``no importance'', so no effect on the output. $CU=1$ signifies that the input value(s) are the best possible for obtaining a high-utility output value and $CU=0$ signifies the worst possible input value(s). The value range for Contextual influence is $[-\phi_{0},-\phi_{0}+1]$, so Contextual influence can also be interpreted directly. Shapley value and LIME don't have such pre-defined ranges, which may lead to misinterpretations of $\phi$ values, especially when $\phi$ values are close to zero. 

CIU (and therefore also Contextual influence) values have no or small variance in the experimental results, whereas Shapley value and LIME show considerable variance on consecutive runs for the same instance. Therefore, CIU and Contextual influence explanations remain identical over consecutive runs, whereas the Shapley value and LIME results change from one run to the next. Such variance is a challenge for the trustworthiness of the explanations produced by Shapley value and LIME and could be a major problem in real-world use cases, such as explaining credit worthiness assessments given by AI systems. 

Contextual importance corresponds to the intuitive and common definition of importance used in decision theory, at least for linear models, and can therefore be generalised to global feature importance. This is not the case for influence values ($\phi$) that change for every instance $x$ even in the linear case. Therefore it seems reasonable to use CI for estimating global importance also in the non-linear case, rather than using $mean(|\phi|)$ values. 

The theory and results suggest that CIU could provide more informative and stable explanations than the studied mainstream methods. This paper focuses on so called ``tabular data'' data sets but there's no reason for why CIU wouldn't be applicable to other kinds of data, such as images, text \etc, which have been extensively studied for Shapley value and LIME.

\newpage

\bibliographystyle{splncs04}
\bibliography{Bib.bib}

\end{document}